\documentclass[sigconf]{acmart}
\usepackage{amsmath}
\usepackage{amssymb}
\usepackage{amsthm,multirow}
\usepackage{mathtools}
\usepackage{booktabs}
\usepackage{array,enumitem}
\usepackage[linesnumbered,ruled]{algorithm2e}
\usepackage{bm,changepage}
\usepackage{subcaption,siunitx,booktabs}
\newcommand{\thickbar}[1]{\mathbf{\bar{\text{$#1$}}}}
\AtBeginDocument{%
  \providecommand\BibTeX{{%
    \normalfont B\kern-0.5em{\scshape i\kern-0.25em b}\kern-0.8em\TeX}}}

\newcommand{\proposed}{\textsf{asp2vec}}
\newcommand{\proposedhet}{\textsf{asp2vec-het}}
\newcommand{\etal}{\textit{et al}. }
\copyrightyear{2020}
\acmYear{2020}
\setcopyright{acmcopyright}
\acmConference[KDD '20] {26th ACM SIGKDD Conference on Knowledge Discovery and Data Mining}{August 23--27, 2020}{Virtual Event, USA}
\acmBooktitle{26th ACM SIGKDD Conference on Knowledge Discovery and Data Mining (KDD '20), August 23--27, 2020, Virtual Event, USA}
\acmPrice{15.00}
\acmDOI{10.1145/3394486.3403196}
\acmISBN{978-1-4503-7998-4/20/08}

\begin{document}
\fancyhead{}
\title{Unsupervised Differentiable Multi-aspect Network Embedding}

\author{Chanyoung Park$^{1}$, Carl Yang$^{2}$, Qi Zhu$^{1}$, Donghyun Kim$^{4}$, Hwanjo Yu$^{3*}$, Jiawei Han$^{1}$}
\thanks{*Corresponding Author}
\affiliation{%
	\institution{$^{1}$Department of Computer Science, University of Illinois at Urbana-Champaign, IL, USA}
}
\affiliation{%
	\institution{$^{2}$Emory University, GA, USA, $^{3}$POSTECH, Pohang, South Korea, $^{4}$Yahoo! Research, CA, USA}
}
\email{{pcy1302, qiz3, hanj}@illinois.edu,j.carlyang@emory.edu,donghyun.kim@verizonmedia.com, hwanjoyu@postech.ac.kr}

\begin{abstract}
Network embedding is an influential graph mining technique
for representing nodes in a graph as distributed vectors. However, the majority of network embedding methods focus on learning a single vector representation for each node, which has been recently criticized for not being capable of modeling multiple aspects of a node. To capture the multiple aspects of each node, existing studies mainly rely on offline graph clustering performed prior to the actual embedding, which results in the cluster membership of each node (i.e., node aspect distribution) fixed throughout training of the embedding model.
We argue that this not only makes each node always have the same aspect distribution regardless of its dynamic context, but also hinders the end-to-end training of the model that  
eventually leads to the final embedding quality largely dependent on the clustering.
In this paper, we propose a novel end-to-end framework for {m}ulti-{a}spect {n}etwork {em}bedding, called~\proposed, in which the aspects of each node are dynamically assigned based on its local context. More precisely, among multiple aspects, we dynamically assign a \textit{single} aspect to each node based on its current context, and our aspect selection module is end-to-end differentiable via the Gumbel-Softmax trick. We also introduce the aspect regularization framework to capture the interactions among the multiple aspects in terms of relatedness and diversity. We further demonstrate that our proposed framework can be readily extended to heterogeneous networks. Extensive experiments towards various downstream tasks on various types of homogeneous networks and a heterogeneous network demonstrate the superiority of~\proposed. 
\end{abstract}

\ccsdesc[300]{Computing methodologies~Unsupervised learning}

\keywords{Network Embedding, Representation Learning, Graph Mining}

\maketitle

\section{Introduction}
Networks constitute a natural paradigm to represent real-world relational data that contain various relationships between entities ranging from online social network of users, and academic publication network of authors, to protein-protein interaction (PPI) network in the physical world. Due to the pervasive nature of networks, analyzing and mining useful knowledge from networks has been an actively researched topic for the past decades.
Among various tools for network analysis, \textit{network embedding}, which learns continuous vector representations for nodes in a network, has recently garnered attention, and has been effectively and efficiently applied to various downstream network-based applications, such as node classification~\cite{dong2017metapath2vec,park2019unsupervised,liu2019single}, and link prediction~\cite{abu2017learning,epasto2019single}.

\begin{figure}
	\centering
	\includegraphics[width=0.75\linewidth]{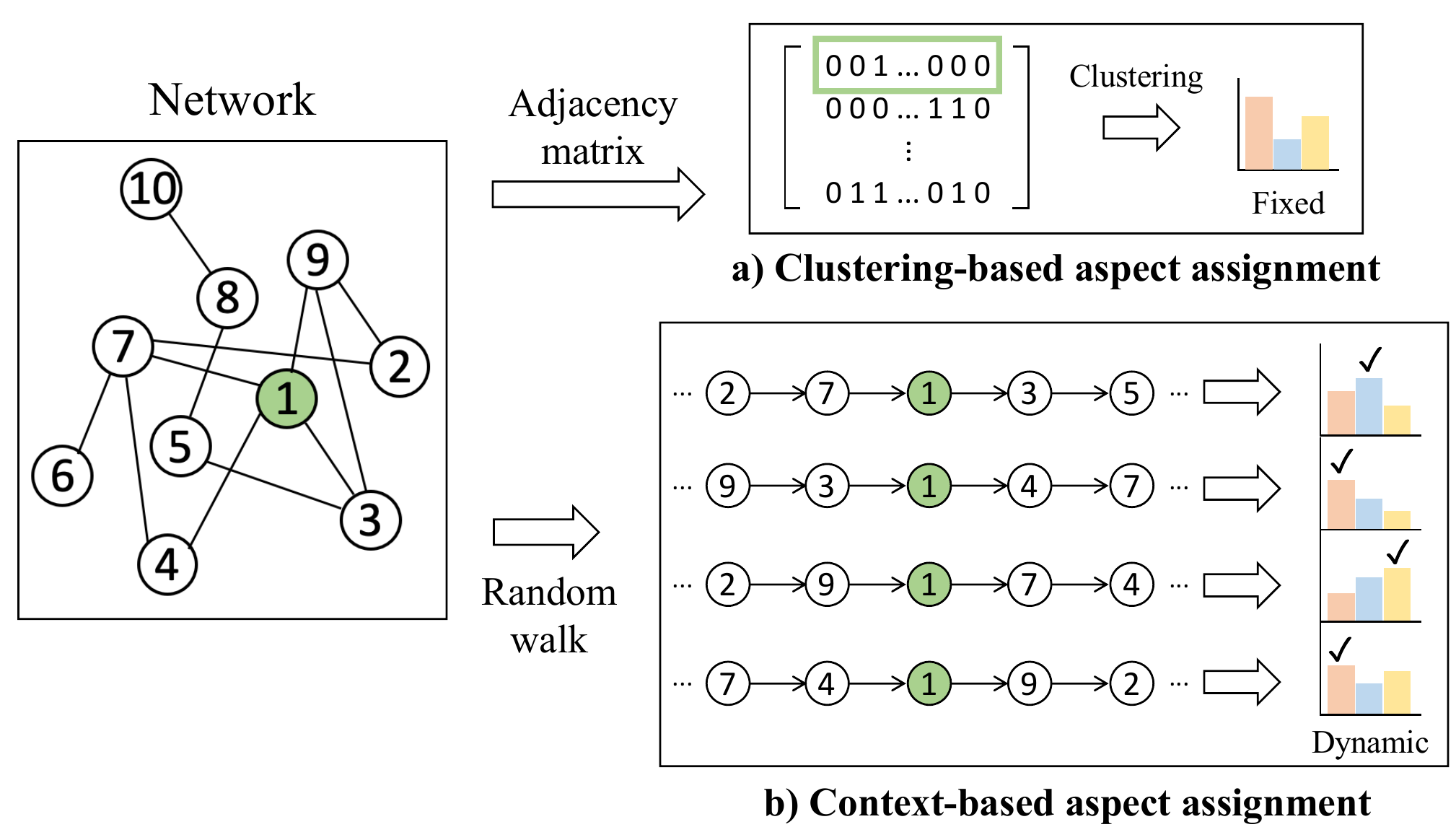}
	\vspace{-1ex}
	\caption{a) Clustering--based aspect assignment that fixes the aspect distribution during the embedding learning. b)~\proposed~dynamically selects a single aspect based on the local context nodes. }
	\label{fig:toy_1}	
	\vspace{-2ex}
\end{figure}

A common underlying idea of network embedding methods is that a node embedding vector should be able to preserve the neighborhood structure of the node, i.e., local structural characteristics. Deepwalk~\cite{perozzi2014deepwalk} is a pioneering method that leverages the node co-occurrence information to learn the representations of nodes in a network such that nodes that frequently co-occur together have similar vector representations~\cite{grover2016node2vec}. Specifically, the co-occurrence based methods usually perform random walks on a network to obtain node sequences on which the skip-gram~\cite{mikolov2013distributed} model is applied.
However, past research on network embedding mostly assumes the existence of a single vector representation for each node, whereas in reality a node usually has multiple aspects. For example, nodes (e.g., authors) in an academic publication network often belong to multiple research communities, and thus modeling each node with a single vector entails information loss. 

In this regard, several attempts have been recently made to model the multiple aspects of a node by learning multiple vector representations for each node~\cite{epasto2019single,liu2019single,wang2019mcne}. However, there still remain several limitations.
First, the aspects of the nodes are determined based on an offline clustering algorithm performed in advance (Figure~\ref{fig:toy_1}a).
In particular, to determine the node aspects, PolyDW~\cite{liu2019single} performs clustering based on matrix factorization (MF) on the adjacency matrix, and similarly, Splitter~\cite{epasto2019single} performs a local clustering algorithm based on ego-network analysis that encodes the role of a node in different local communities. However, as the clustering is done prior to the actual embedding learning, the cluster membership distribution (i.e., aspect distribution) for each node is fixed, which leads to each node always having the same aspect distribution regardless of its current context. This is especially problematic for the co-occurrence based network embedding methods, because the context of a node changes dynamically over multiple random walks. Moreover, owing to the offline clustering, the embedding methods cannot be trained in an end-to-end manner, which eventually leads to the final embedding quality largely dependent on the clustering.

Another limitation is that the interactions among the aspects are not explicitly captured. Although the interactions among aspects are implicitly revealed by the cluster membership of each node, it is only used for sampling the node aspects~\cite{epasto2019single}, and not directly incorporated for training the aspect embedding vectors.
Going back to our example of academic publication network, authors can belong to multiple research communities, and these communities interact with each other. For example, data mining (DM) and database (DB) communities are more related to each other than data mining and computer architecture (CA) communities, and thus such interactions (i.e., relatedness: DM $\leftrightarrow$ DB, and diversity: DM $\leftrightarrow$ CA) should be captured. Furthermore, viewing the aspects in a broader perspective, we argue that the interactions among aspects differ according to the inherent characteristics of the networks. For example, nodes in an academic publication network tend to belong to fewer number of communities (i.e., aspects are inherently less diverse) than nodes in a PPI network. This is because an ordinary author in an academic publication network usually works on a limited number of research topics, whereas each protein in a PPI network is associated with various functions and involved in various interactions.
As such, modeling the interactions among aspects is challenging, and thus should be done in a systematic way rather than implicitly by clustering.

In the light of these issues, we propose~\proposed, a novel end-to-end framework for {m}ulti-{a}spect {n}etwork {em}bedding. The core idea is to dynamically assign aspects to each node according to its local context as illustrated in Figure~\ref{fig:toy_1}b (\textbf{Sec.~\ref{sec:cbmanr}}). More precisely, we selectively assign (sample) a \textit{single} aspect for each node based on our assumption that each node should belong to a single aspect under a certain \textit{local context}. For example, even an author with diverse research areas, who belongs to multiple research communities (i.e., multiple aspects), focuses on a single research topic  when
collaborating with a particular group of people (i.e., a single aspect within a local context).
We materialize this idea by devising the \textit{aspect selection module} based on the Gumbel-Softmax trick~\cite{jang2016categorical}, which approximates the sampling from a categorical distribution in a differentiable fashion, thereby enabling an end-to-end training procedure (\textbf{Sec.~\ref{sec:gumbel}}).
Moreover, we introduce the \textit{aspect regularization framework} to simultaneously capture the interactions and relationships among aspects in terms of both relatedness and diversity (\textbf{Sec.~\ref{sec:reg}}).
Finally, we demonstrate that our proposed framework can be readily extended to heterogeneous networks whose nodes and edges are multiple-typed (\textbf{Sec.~\ref{sec:het}}).
Through our extensive experiments on 13 real-world datasets, including various types of homogeneous networks and a heterogeneous network, we demonstrate the effectiveness of~\proposed~in multiple downstream tasks including link prediction and author identification, in comparison with state-of-the-art multi-aspect network embedding methods. 
The source code of~\proposed~can be found in https://github.com/pcy1302/asp2vec/.

\section{Related Work}
\noindent{\textbf{Network embedding}. }
Network embedding methods aim at learning low-dimensional vector representations for nodes in a graph while preserving the network structure~\cite{perozzi2014deepwalk,grover2016node2vec}, and various other properties such as node attributes~\cite{park2019unsupervised}, and structural role~\cite{ribeiro2017struc2vec}.
More precisely, inspired by word2vec~\cite{mikolov2013distributed}, DeepWalk~\cite{perozzi2014deepwalk} and node2vec~\cite{grover2016node2vec} perform truncated random walks on graphs to obtain multiple node sequences, and then perform skip-gram to learn node embeddings by making an analogy between random walk sequences on a network and sentences in natural language.
In another line of research, graph neural networks (GNNs)~\cite{wu2019comprehensive} recently have drawn
intensive attention. Their main idea is to represent a node by aggregating information from its neighborhood ~\cite{kipf2016semi,velivckovic2017graph}. 
However, their performance largely depends on the available node labels~\cite{hamilton2017inductive,kipf2016semi,ma2019disentangled}, whereas our proposed framework is fully unsupervised.
Recently, Velivckovic et al~\cite{velivckovic2018deep} proposed a GNN--based unsupervised network embedding method, called DGI, which maximizes the mutual information between the global representation of a graph and its local patches, i.e., node embeddings. 
However, while random walks naturally capture higher-order structural information and allow for the characterizing of different node contexts in networks, the performance of GNNs degenerates with larger number of convolution layers~\cite{li2018deeper}.
Moreover, unlike random walk--based algorithms~\cite{dong2017metapath2vec}, GNNs cannot be easily applied to heterogeneous networks without complicated design~\cite{schlichtkrull2018modeling,park2019unsupervised}.

\smallskip
\noindent\textbf{Multi-aspect network embedding. }
While most network embedding methods focus on learning a single embedding vector for each node, several recent methods~\cite{epasto2019single,liu2019single,wang2019mcne,yang2018multi,sun2019vgraph,ma2019disentangled} have been proposed to learn multiple embedding vectors for each node in a graph. More precisely, PolyDW~\cite{liu2019single} performs MF--based clustering to determine the aspect distribution for each node. Then, given random walk sequences on a graph, the aspect of the target and the context nodes are sampled independently according to the aspect distribution, and thus the target and the context nodes may have different aspects. 
Moreover, although capturing the local neighborhood structure~\cite{yang2018multi} is more important for the co-occurrence based network embedding methods, PolyDW exploits clustering that focuses on capturing the global structure.
Splitter~\cite{epasto2019single} performs local clustering to split each node into multi-aspect representations, however, it blindly trains all the aspect embedding vectors to be close to the original node embedding vector rather than considering the aspect of the target node. 
MCNE~\cite{wang2019mcne} introduces a binary mask layer to split a single vector into multiple conditional embeddings, however, the number of aspects should be defined in the datasets (e.g., number of different types of user behaviors). Moreover, it formalizes the task of item recommendation as a supervised task, and combines network embedding methods with BPR~\cite{rendle2012bpr}, whereas our proposed framework is trained in a fully unsupervised manner with an arbitrary number of aspects. MNE~\cite{yang2018multi} trains multi-aspect node embeddings based on MF, while considering the diversity of the aspect embeddings. However, the aspect selection process is ignored, and the aspect embeddings are simply trained to be as dissimilar as possible to each other without preserving any relatedness among them. 

\smallskip
\noindent\textbf{Heterogeneous network embedding. }
A heterogeneous network (HetNet) contains multi-typed nodes and multi-typed edges, which should be modeled differently from the homogeneous counterpart, and there has been a line of research on heterogeneous network embedding~\cite{fu2017hin2vec,dong2017metapath2vec,hussein2018meta}.
Specifically, metapath2vec~\cite{dong2017metapath2vec} extends Deepwalk by introducing a random walk scheme that is conditioned on meta-paths, where the node embeddings are learned by heterogeneous skip-gram. 
While various single aspect embedding methods have been proposed, multi-aspect network embedding is still in its infancy. More precisely, a line of work~\cite{shi2018easing,chen2018pme} commonly projects nodes into multiple embedding spaces defined by multiple aspects, and the node embeddings are trained on each space.  However, these methods define the aspects based on the edges whose ground-truth labels should be predefined (e.g., the edge between ``Obama'' and ``Honolulu'' is labeled as ``was\_born\_in''), which is not always the case in reality. Finally, TaPEm~\cite{park2019task} introduces the pair embedding framework that directly models the relationship between two nodes of different types, but still each node has a single embedding vector.
Furthermore, although GNNs have also been recently applied to HetNets, they are either semi-supervised methods~\cite{wang2019heterogeneous}, or cannot be applied to a genuine HetNet whose both nodes and edges are multiple typed~\cite{park2019unsupervised}.


\section{Preliminary} 
\noindent{\textbf{Random Walk--based Unsupervised Network Embedding. }}
\label{sec:DW}
Inspired by the recent advancement of deep learning methods, especially, skip-gram based word2vec~\cite{mikolov2013distributed}, previous network representation learning methods~\cite{grover2016node2vec,perozzi2014deepwalk} started viewing nodes in a graph as words in a document. More precisely, these methods commonly perform a set of truncated random walks $\mathcal{W}$ on a graph, which are analogous to sequences of words, i.e., sentences. Then, the representation for each node is learned by optimizing the skip-gram objective, which is to maximize the likelihood of the context given the target node:
\begin{equation}
\small
\begin{split}
\max\sum_{\mathbf{w}\in\mathcal{W}}\sum_{v_i\in \mathbf{w}}\sum_{v_j\in \mathcal{N}(v_i)}\log\frac{\exp(\langle \mathbf{P}_i,\mathbf{Q}_j\rangle)}{\sum_{v_j{^\prime}\in \mathcal{V}}\exp(\langle \mathbf{P}_i,\mathbf{Q}_{j{^\prime}}\rangle)}
\end{split}
\label{eqn:DW}
\end{equation}
where $\textbf{P}_i\in\mathbb{R}^d$ is the target embedding vector for node $v_i$, $\textbf{Q}_j\in\mathbb{R}^d$ is the context embedding vector for node $v_j$, $d$ is the number of embedding dimensions, $\langle a,b \rangle$ denotes a dot product between $a$ and $b$, $\textbf{w}$ is a node sequence, and $\mathcal{N}(v_i)$ denotes the nodes within the context window of node $v_i$. The final embedding vector for node $v_i$ is generally obtained by averaging its target and context embedding vector. i.e., $(\mathbf{P}_i+\mathbf{Q}_i)/2$. By maximizing the above Eqn.~\ref{eqn:DW}, nodes that frequently appear together within a context window will be trained to have similar embeddings. Despite its effectiveness in learning node representations, it inherently suffers from a limitation that \textit{each node is represented by a single embedding vector}. 

\subsection{Problem Statement}
\theoremstyle{definition}
\begin{definition}{(\textbf{Multi-aspect Node Representation Learning})}\\
	Given a graph $\mathcal{G}=(\mathcal{V},\mathcal{E})$, where $\mathcal{V}=\{v_1,v_2,...,v_n\}$ represents the set of nodes, and $\mathcal{E}=\{e_{i,j}\}_{i,j}^n$ represents the set of edges, where $e_{i,j}$ connects $v_i$ and $v_j$, we aim to derive a multi-aspect embedding matrix $\mathbf{Q}_i\in\mathbb{R}^{K\times d}$ for each node $v_i$, where $K$ is an arbitrary predefined number of aspects. More precisely, each aspect embedding vector $\{\mathbf{Q}_i^{(s)}\in\mathbb{R}^d\}_{s=1}^K$ should 1) preserve the network structure information, 2) capture node $v_i$'s different aspects, and 3) model the interactions among the aspects (i.e., relatedness and diversity).
\end{definition}

%
\section{Proposed Framework:~\proposed}
\begin{figure*}
	\centering
	\includegraphics[width=0.75\linewidth]{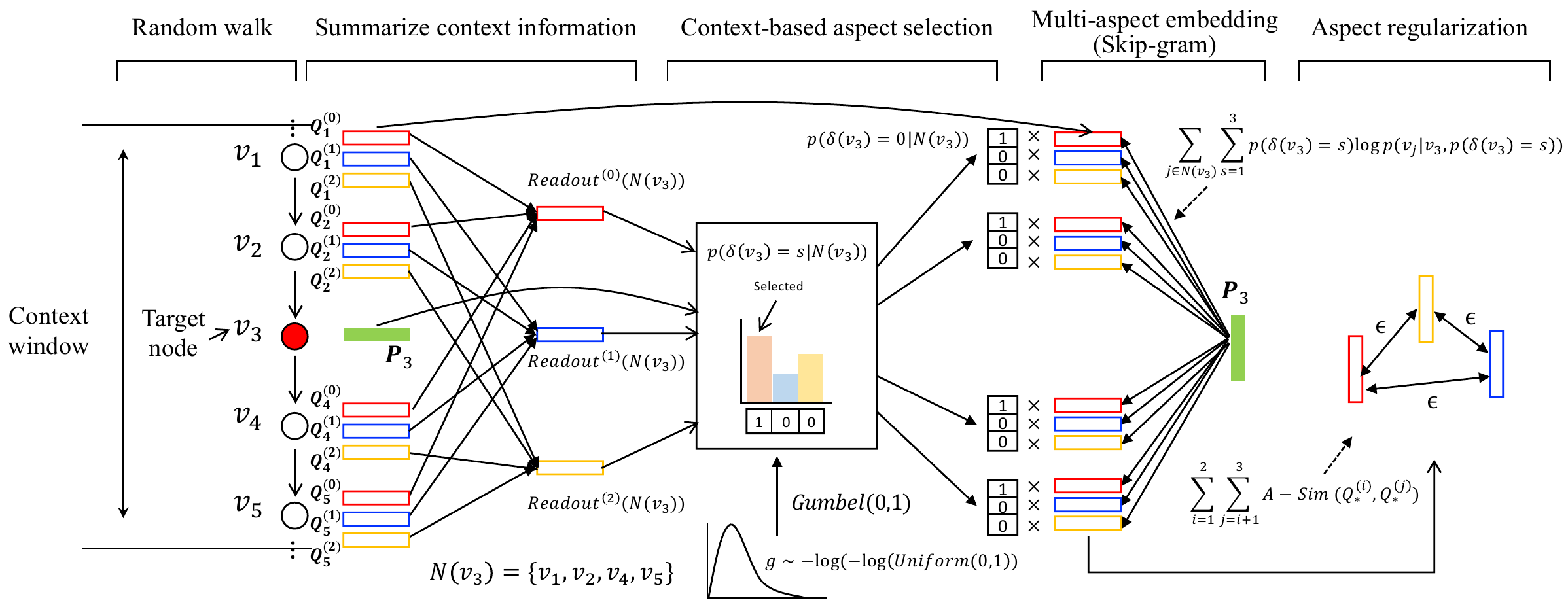}
	\vspace{-2ex}
	\caption{The overall architecture of~\proposed. Given a sequence of nodes within a context window, a single aspect is dynamically selected according to the aspect of the context nodes. Then, the skip-gram is performed based on the selected aspect. The aspect regularization framework encourages the diversity and preserves the relatedness among the aspect embedding vectors.}
	\label{fig:overview}	
	\vspace{-2ex}
\end{figure*}

We present our proposed framework for context--based {m}ulti--{a}spect {n}etwork {em}bedding (Sec.~\ref{sec:cbmanr}) that includes the aspect selection module (Sec.~\ref{sec:gumbel}) and the aspect regularization framework (Sec.~\ref{sec:reg}). Then, we demonstrate how~\proposed~can be further extended to a heterogeneous network (Sec~\ref{sec:het}). Figure~\ref{fig:overview} summarizes the overall architecture of our proposed framework, called~\proposed.


\subsection{Context--based Multi--aspect Network Embedding}
\label{sec:cbmanr}
Given a target node and its local context, the core idea is to first determine the current aspect of the target node based on its local context, and then predict the context node with respect to the selected aspect of the target node. More precisely, given a target node embedding vector $\mathbf{P}_i\in\mathbb{R}^d$, and its currently selected aspect $\delta({v_i})\in\{1,2,...,K\}$, our goal is to predict its context embedding vectors with respect to the selected aspect $\delta({v_i})$. i.e., $\{\mathbf{Q}_j^{(\delta({v_i}))} | v_j\in\mathcal{N}(v_i)\}$. Formally, for each random walk sequence $\mathbf{w}\in\mathcal{W}$, we aim to maximize the following objective:

\begin{equation}
\small
\begin{split}
\mathcal{J}_{\proposed}^{(\mathbf{w})}&=\sum_{v_i\in \mathbf{w}}\sum_{v_j\in \mathcal{N}(v_i)}\sum_{s=1}^{K}p(\delta(v_i)=s)\log p(v_j | v_i, p(\delta(v_i)=s))\\
&=\sum_{v_i\in \mathbf{w}}\sum_{v_j\in \mathcal{N}(v_i)}\sum_{s=1}^{K}p(\delta(v_i)=s)\log\frac{\exp(\langle \mathbf{P}_i,\mathbf{Q}^{(s)}_j\rangle)}{\sum_{v_{j^\prime}\in \mathcal{V}}\exp(\langle \mathbf{P}_i,\mathbf{Q}^{(s)}_{j{^\prime}}\rangle)}
\end{split}
\label{eqn:walk_loss}
\raisetag{35pt}
\end{equation}
where $\mathbf{Q}_j^{(s)}\in\mathbb{R}^d$ is the embedding vector of node $v_j$ in terms of aspect $s$, $p(\delta(v_i)=s)$ denotes the probability of $v_i$ being selected to belong to the aspect $s$, where $\sum_{s=1}^{K}p(\delta(v_i)=s)=1$, and $p(v_j | v_i, p(\delta(v_i)=s))$ denotes the probability of a context node $v_j$ given a target node $v_i$ whose aspect is $s$.
Note that this is in contrast to directly maximizing the probability of its context node regardless of the aspect of the target node as in Eqn.~\ref{eqn:DW}. 

\subsection{Determining the Aspect of the Center Node}
\label{sec:gumbel}
Next, we introduce the \textit{aspect selection module} that computes the aspect selection probability $p(\delta(v_i))$. Here, we assume that the aspect of each target node $v_i$ can be readily determined by examining its local context $\mathcal{N}(v_i)$. i.e.,  $p(\delta(v_i))\equiv p(\delta(v_i)|\mathcal{N}(v_i))$. More precisely, we apply softmax to model the probability of $v_i$'s aspect:
\begin{equation}
\small
\begin{split}
p(\delta(v_i)=s)&\equiv p(\delta(v_i)=s|\mathcal{N}(v_i))=\textsf{softmax}(\langle \mathbf{P}_i,\textsf{Readout}^{(s)}(\mathcal{N}(v_i))\rangle)\\
&=\frac{\exp(\langle \mathbf{P}_i,\textsf{Readout}^{(s)}(\mathcal{N}(v_i))\rangle)}{\sum_{s^\prime=1}^{K}\exp(\langle \mathbf{P}_i,\textsf{Readout}^{(s^{\prime})}(\mathcal{N}(v_i))\rangle)}
\end{split}
\label{eqn:SM}
\raisetag{20pt}
\end{equation}
where we leverage a readout function $\textsf{Readout}^{(s)}:\mathbb{R}^{|\mathcal{N}(v_i)|\times d}$$\rightarrow$$\mathbb{R}^d$ to summarize the information encoded in the local context of node $v_i$, i.e., $\mathcal{N}(v_i)$, with respect to aspect $s$. 

\subsubsection{\textbf{Gumbel-Softmax based Aspect Selection} }
Note that modeling the aspect selection probability $p(\delta(v_i))$ based on softmax (Eqn.~\ref{eqn:SM}) assigns a probability distribution over the number of aspects $K$, which is \textit{continuous}. However, we argue that although a node may  simultaneously belong to multiple aspects in a global perspective, \textit{it should be assigned to a single aspect within a local context.} 
For example, even an author with diverse research areas, who belongs to multiple research communities, focuses on a single research topic when collaborating with a particular group of people.
In other words, the aspect selection should be done in a discrete manner based on a categorical distribution over $K$ aspects. i.e., hard selection. 
However, since the hard selection is a non-differentiable operation, it blocks the flow of the gradients down to the embedding vectors, and prevents the end-to-end back-propagation optimization. As a workaround, we adopt the Gumbel-Softmax trick~\cite{jang2016categorical}, which approximates the sampling from a categorical distribution by reparameterizing the gradients with respect to the parameters of the distribution.
More precisely, given a $K$-dimensional categorical distribution with class probability $\pi_1,\pi_2,..,\pi_K$, the gumbel-max trick provides a simple way to draw a one-hot sample $z=\left(z_{1}, \cdots z_{K}\right) \in \mathbb{R}^{K}$ from the categorical distribution as follows:
\begin{equation}
{z}=\textsf{one-hot}\left(\textsf{argmax}_{i}\left[\log \pi_{i}+g_{i}\right]\right)
\label{eqn:onehot}
\end{equation}
where $g_i$ is the gumbel noise drawn from Gumbel(0,1) distribution, which can be obtained as follows:
\begin{equation}
g_i=-\log(-\log(u_i)) \qquad u_i \sim Uniform(0,1)
\end{equation}
Due to the non-differentiable nature of the \textsf{argmax($\cdot$)} operation in Eqn.~\ref{eqn:onehot}, we further approximate it by using softmax to ensure the differentiability as follows, which is known as the Gumbel-Softmax trick~\cite{jang2016categorical}:
\begin{equation}
\small
\begin{split}
z_i &= \textsf{softmax}\left[\log \pi_{i}+g_{i}\right] \\
&=\frac{\exp \left(\left(\log \pi_{i}+g_{i}\right) / \tau\right)}{\sum_{j=1}^{K} \exp \left(\left(\log \pi_{j}+g_{j}\right) / \tau\right)} \quad \text { for } k=1, \ldots, K
\end{split}
\label{eqn:GSM}
\end{equation}
where $\tau$ is the temperature parameter to control the extent to which the output approximates the \textsf{argmax($\cdot$)} operation: As $\tau\rightarrow0$, samples from the Gumbel-Softmax distribution become one-hot. Finally, replacing the softmax in Eqn.~\ref{eqn:SM} with the Gumbel-Softmax in Eqn.~\ref{eqn:GSM}, we obtain the following aspect selection probability:
\begin{equation}
\small
p(\delta(v_i)=s|\mathcal{N}(v_i))=\frac{\operatorname{exp}[(\log{\langle \mathbf{P}_i,\textsf{Readout}^{(s)}(\mathcal{N}(v_i))\rangle} + g_s) / \tau]}{\sum_{s^\prime=1}^{K}\operatorname{exp}[(\log{\langle \mathbf{P}_i,\textsf{Readout}^{(s^{\prime})}(\mathcal{N}(v_i))\rangle} + g_{s^\prime})/\tau]}
\label{eqn:gumbel}
\end{equation}
Combining the loss for all walks $\mathbf{w}\in\mathcal{W}$ described in Eqn.~\ref{eqn:walk_loss}, the minimization objective for learning the context-based multi-aspect node representation is given by:
\begin{equation}
\small
\mathcal{L}_\proposed=-\sum_{\mathbf{w}\in\mathcal{W}}\mathcal{J}_{\proposed}^{(\mathbf{w})}
\label{eqn:maDW}
\end{equation}


\subsubsection{\textbf{Readout Function}}
\label{sec:readout}
Recall that we expect the output of $\textsf{Readout}^{(s)}(\mathcal{N}(v_i))$ to capture the current aspect $s$ of the target node $v_i$ by leveraging its local context information $\mathcal{N}(v_i)$. Among various choices for $\textsf{Readout}(\cdot)$, we choose the average pooling operation:
\begin{equation}
\small
 \textsf{Readout}^{(s)}(\mathcal{N}(v_i)) =  \frac{1}{|\mathcal{N}(v_i)|}\sum_{v_{j}\in \mathcal{N}(v_i)}\mathbf{Q}_j^{(s)}=\thickbar{\mathbf{Q}}_{\mathcal{N}(v_i)}^{(s)} 
\end{equation}
It is worth noting that thanks to its simplicity and efficiency, we choose the average pooling over the more advanced techniques that assign different weights to each context node, such as recurrent neural network--based or attention--based pooling~\cite{zhou2016attention}.

\smallskip
\noindent\textbf{Discussion. } If each node is associated with some attribute information, determining the aspect of each node can be more intuitive than solely relying on the context node embeddings originally learned to preserve the graph structure. In this regard, we study in Sec.~\ref{sec:aspect_het} how to model the aspect when node attributes are given.
\subsection{Modeling Relationship Among Aspects}
\label{sec:reg}
As different aspect embeddings are intended to capture different semantics of each node, they should primarily be sufficiently diverse among themselves. However, at the same time,
these \textit{aspects are inherently related to each other} to some extent. For example, in an academic publication network, each node (i.e., author) may belong to various research communities (i.e.aspects), such as DM, CA, and DB community. In this example scenario, we can easily tell that DM and DB communities are more related than DM and CA communities, which shows the importance of modeling the interaction among aspects. In short, aspect embeddings should not only be 1) \textbf{diverse} (e.g., DM$\leftrightarrow$CA) so as to independently capture the inherent properties of individual aspects, but also 2) \textbf{related} (e.g., DM$\leftrightarrow$DB) to each other to some extent so as to capture some common information shared among aspects.

To this end, we introduce a novel aspect regularization framework, called $\text{reg}_{\textsf{asp}}$, which is given by:
\begin{equation}
\small
\text{reg}_{\textsf{asp}}=\sum_{i=1}^{K-1}\sum_{j=i+1}^{K}\operatorname{A-Sim}(\mathbf{Q}_*^{(i)},\mathbf{Q}_*^{(j)})
\label{eqn:reg}
\end{equation}
where $\mathbf{Q}_*^{(i)}\in\mathbb{R}^{n\times d}$ is the aspect embedding matrix in terms of aspect $i$ for all nodes $v\in\mathcal{V}$; $|\mathcal{V}|=n$, and $\operatorname{A-Sim}(\cdot,\cdot)$ measures the similarity score between two aspect embedding matrices: large $\operatorname{A-Sim}(\mathbf{Q}_*^{(i)},\mathbf{Q}_*^{(j)})$ refers to the two aspects sharing some information in common, whereas small $\operatorname{A-Sim}(\mathbf{Q}_*^{(i)},\mathbf{Q}_*^{(j)})$ means aspect $i$ and aspect $j$ capture vastly different properties.
The aspect similarity score $\operatorname{A-Sim}(\mathbf{Q}_*^{(i)},\mathbf{Q}_*^{(j)})$ is computed by the sum of the similarity scores between aspect embeddings of each node:
\begin{equation}
\small
\operatorname{A-Sim}(\mathbf{Q}_*^{(i)},\mathbf{Q}_*^{(j)})=\sum_{h=1}^{|V|}f(\mathbf{Q}_h^{(i)},\mathbf{Q}_h^{(j)})
\label{eqn:adiv}
\end{equation}
where $\mathbf{Q}_h^{(i)}\in\mathbb{R}^{d}$ is the embedding vector of node $v_h$ in terms of aspect $i$, and $f(\mathbf{Q}_h^{(i)},\mathbf{Q}_h^{(j)})$ denotes the aspect similarity score (ASS) between embeddings of aspects $i$ and $j$ with respect to node $v_h$. 
By minimizing $\text{reg}_{\textsf{asp}}$ in Eqn.~\ref{eqn:reg}, we aim to learn \textit{diverse} aspect embeddings capturing inherent properties of each aspect.
In this work, we evaluate the ASS based on the cosine similarity between two aspect embedding vectors given by the following equation:
\begin{equation}
\small
f(\mathbf{Q}_h^{(i)},\mathbf{Q}_h^{(j)})
= \frac{\langle \mathbf{Q}_h^{(i)},\mathbf{Q}_h^{(j)} \rangle}{\lVert \mathbf{Q}_h^{(i)}\rVert\lVert \mathbf{Q}_h^{(j)}\rVert}, \quad -1\leq f(\mathbf{Q}_h^{(i)},\mathbf{Q}_h^{(j)}) \leq 1
\label{eqn:MNAR}
\end{equation}

However, as mentioned earlier, the aspect embeddings should not only be diverse but also to some extent \textit{related} to each other (e.g., DM$\leftrightarrow$DB). In other words, as the aspects are not completely independent from each other, we should model their interactions. 
To this end, we introduce a binary mask $w_{i,j}^h$ to selectively penalize the aspect embedding pairs according to their ASSs~\cite{ayinde2019regularizing}. More precisely, the binary mask is defined as:
\begin{equation}
\small
	w^h_{i, j}=\left\{\begin{array}{ll}{1,} & {\left|f(\mathbf{Q}_h^{(i)},\mathbf{Q}_h^{(j)})\right| \geq \epsilon} \\ {0,} & {\text { otherwise }}\end{array}\right.
\label{eqn:mask}
\end{equation}
where $\epsilon$ is a threshold parameter that controls the amount of information shared between a pair of aspect embedding vectors of a given node: large $\epsilon$ encourages the aspect embeddings to be related to each other, whereas small $\epsilon$ encourages diverse aspect embeddings. Specifically, for a given node $v_h$, if the absolute value of ASS between the aspect embedding vectors of aspect $i$ and $j$, i.e., $\left|f(\mathbf{Q}_h^{(i)},\mathbf{Q}_h^{(j)})\right|$, is greater than $\epsilon$, then we would like to penalize, and accept it otherwise. In other words, we allow two aspect embedding vectors of a node to be similar to each other to some extent (as much as $\epsilon$). By doing so, we expect that the aspect embeddings to be sufficiently diverse, but at the same time share some information in common. 

Note that the aspect regularization framework is more effective for networks with diverse aspects.
For example, users in social networks tend to belong to multiple (diverse) communities compared with authors in academic networks, because authors usually work on a limited number of research topics. Therefore, as the aspect regularization framework encourages the aspect embeddings to be diverse, it is expected to be more effective in social networks than in academic networks.
We later demonstrate that this is indeed the case, which implies that~\proposed~can also uncover the hidden characteristics of networks.
We consider both positive and negative similarities by taking an absolute value of the cosine similarity, i.e., $\left|f(\cdot,\cdot)\right|$, because negative similarity still means the two aspects are semantically related yet in an opposite direction.

Armed with the binary mask defined in Eqn.~\ref{eqn:mask}, Eqn.~\ref{eqn:adiv} becomes:
\begin{equation}
\small
\operatorname{A-Sim}(\mathbf{Q}_*^{(i)},\mathbf{Q}_*^{(j)})=\sum_{h=1}^{|V|}w^h_{i,j}|f(\mathbf{Q}_h^{(i)},\mathbf{Q}_h^{(j)})|
\end{equation}
The final objective function is to jointly minimize the context--based multi-aspect network embedding loss $\mathcal{L}_\proposed$ in Eqn.~\ref{eqn:maDW}, and the aspect regularization loss $\operatorname{reg}_{\textsf{asp}}$ in Eqn.~\ref{eqn:reg}:
\begin{equation}
\small
\mathcal{L} = \mathcal{L}_{\proposed} + \lambda  \operatorname{reg}_{\textsf{asp}}
\label{eqn:final_loss}
\end{equation}
where $\lambda$ is the coefficient for the aspect regularization.

\smallskip
\noindent\textbf{Discussions on the Number of Parameters. }
A further appeal of our proposed framework is its superior performance with a relatively limited number of parameters. More precisely, PolyDW~\cite{liu2019single}, which is one of the most relevant competitors, requires $2|V|dK$ parameters for node embeddings ($|V|dK$ each for target and context) in addition to the parameters required for the offline clustering method. i.e., matrix factorization, which requires another $2|V|d$, thus $2|V|d(K+1)$ in total. 
Moreover, Splitter~\cite{epasto2019single} requires $|V|d(K+2)$  for node embeddings ($2|V|d $ for Deepwalk parameters and $|V|dK$ for persona embedding), and additional parameters for local and global clustering.
On the other hand, our proposed framework only requires $|V|d(K+1)$ for node embeddings ($|V|d$ for target and $|V|dK$ for context) without additional parameters for clustering. 
Hence, we argue that~\proposed~outperforms state-of-the-art multi-aspect network embedding methods with less parameters. 

\subsection{Task: Link Prediction }
\label{sec:link}
To evaluate the performance of our framework, we perform link prediction, which is to predict the strength of the linkage between two nodes. 
Link prediction is the best choice for evaluating the multi-aspect network embedding methods, because aspect embeddings are originally designed to capture various interactions among nodes, such as membership to multiple communities within a network, which is best revealed by the connection information.
Moreover, link prediction is suggested by recent work as the primary evaluation task for unsupervised network embedding methods compared with node classification that involves a labeling process that may be uncorrelated with the graph itself~\cite{abu2017learning,epasto2019single}.

Recall that in our proposed framework, a node $v_i$ has a center embedding vector $\mathbf{P}_i\in\mathbb{R}^d$, and $K$ different aspect embedding vectors $\{\mathbf{Q}_i^{(s)}\in\mathbb{R}^d\}_{s=1}^K$, which add up to $K+1$ embedding vectors in total for each node. In order to obtain the final embedding vector for node $v_i$, we first compute the average of the aspect embedding vectors, and add it to the center embedding vector:
\begin{equation}
\small
\mathbf{U}_i=\mathbf{P}_i + \frac{1}{K}\sum_{s=1}^{K}\mathbf{Q}_i^{(s)}
\label{eqn:final}
\end{equation}
where $\mathbf{U}_i\in\mathbb{R}^d$ is the final embedding vector for node $v_i$. Note that in previous work~\cite{liu2019single,epasto2019single} that learn multiple embedding vectors for each node, the final link prediction is done by calculating the sum~\cite{liu2019single} or the maximum~\cite{epasto2019single} of the cosine similarity between all possible pairs of aspect embedding vectors. Both require $O(|\mathcal{V}|^2K^2)$ dot product operations to compute the link probability between all pairs of nodes, which is time consuming.
In this work, we simply use the final embedding vector $\mathbf{U}_i$ on which any off-the-shelf classification algorithm, such as logistic regression, is trained facilitating more practical usage in the real-world.

\subsection{Extension to Heterogeneous Network}
\label{sec:het}
Heretofore, we described our approach for learning context--based multi--aspect node representations for a network with a a single type of nodes and a single type of edges. i.e., homogeneous network. 
In this section, we demonstrate that our proposed multi-aspect network embedding framework,~\proposed, can be readily extended to learn representations for nodes in a HetNet. 
Note that PolyDW~\cite{liu2019single} also showed its extendability to a HetNet. However, it is only limited to a bipartite network without node attribute information, whereas~\proposed~can be extended to any type of HetNets with node attribute information.

Recall that in Sec.~\ref{sec:link} we defined our primary task as link prediction. In this regard, among various link prediction tasks that can be defined in a HetNet such as, recommendation~\cite{hu2018leveraging} (i.e., user-item), and author identification (i.e., paper-author), we specifically focus on the task of author identification in big scholarly data whose task is to predict the true authors of an anonymized paper~\cite{chen2017task,park2019task,zhang2018camel}, i.e., link probability between paper-typed nodes and author-typed nodes. Note that each paper is assumed to be associated with its content information. i.e., abstract text.

\begin{figure}
	\centering
	\includegraphics[width=0.9\linewidth]{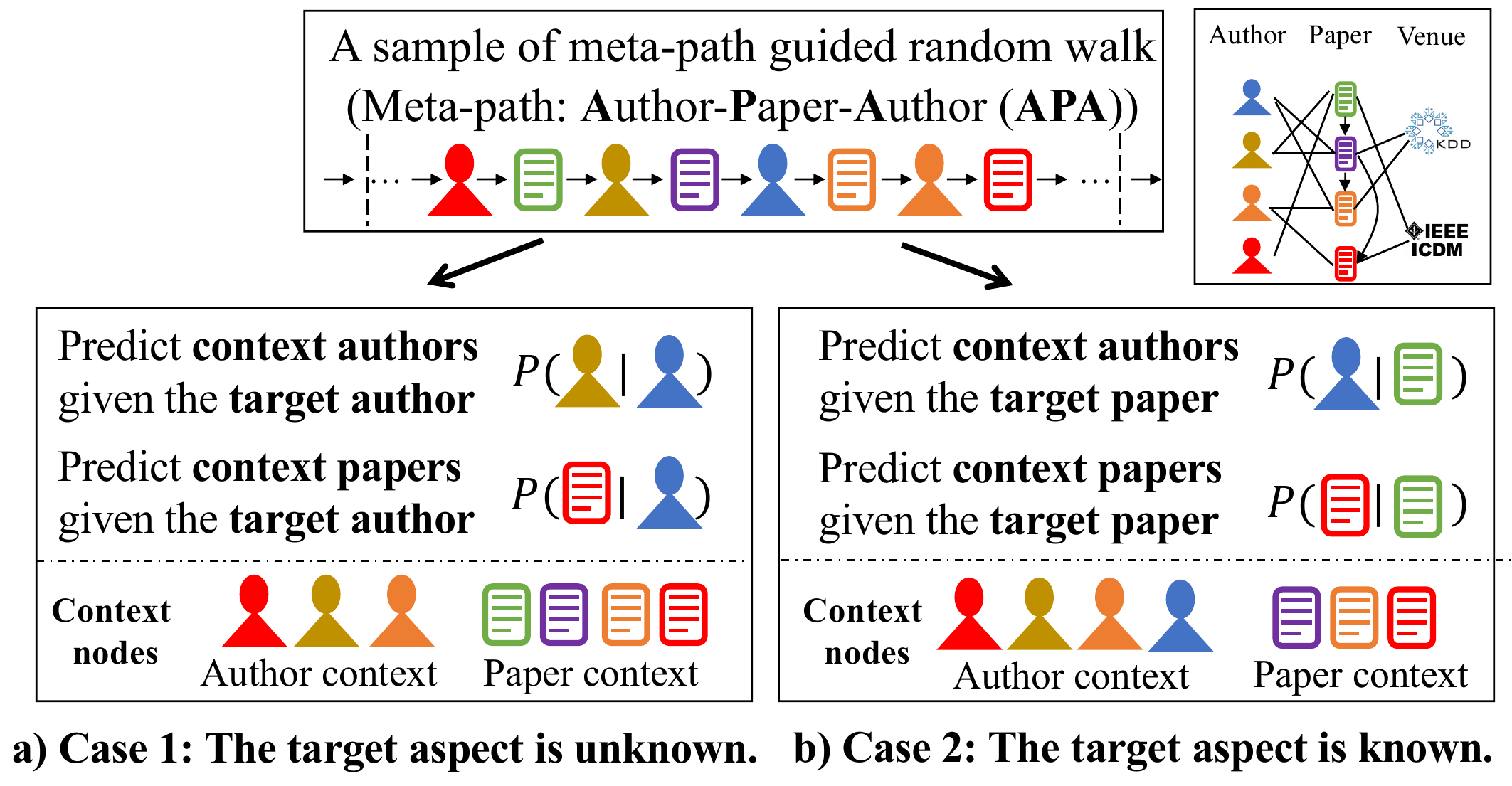}
	\vspace{-2ex}
	\caption{Two different cases should be considered for~\proposedhet.}
	\vspace{-4ex}
	\label{fig:metapath}	
\end{figure}

\subsubsection{\textbf{Context--based Multi--aspect Heterogeneous Network Embedding}}
\label{sec:het_aspect}
To begin with, we first perform meta-path guided random walk~\cite{dong2017metapath2vec} to generate sequences of nodes, which is an extension of the truncated random walk~\cite{perozzi2014deepwalk,grover2016node2vec} to a HetNet. More precisely, random walks are conditioned on a meta-path $\mathcal{P}$~\cite{dong2017metapath2vec}, which is a composition of relations that contains a certain semantic. Figure~\ref{fig:metapath} shows an example metapath ``Author-Paper-Author (APA)'' that refers to a co-authorship relation. After generating a set of random walks $\mathcal{W}_\mathcal{P}$ guided by a meta-path $\mathcal{P}$, we aim to maximize the probability of context nodes given a target node similar to Eqn.~\ref{eqn:walk_loss}. However, unlike the former case, a walk $\mathbf{w}\in\mathcal{W}_{\mathcal{P}}$ now contains multiple types of nodes. Hence, we propose~\proposedhet~by revising Eqn.~\ref{eqn:walk_loss} to incorporate the heterogeneity as follows:
\begin{equation}
\small
\begin{split}
\mathcal{J}^{(\mathbf{w})}_{\proposedhet}=\sum_{v_i \in \mathbf{w}} \sum_{t \in \mathcal{T}_{\mathcal{V}}} \sum_{v_{j} \in \mathcal{N}_{t}(v_i)}\sum_{s=1}^{K}p(\delta(v_i)=s)\log p(v_j | v_i, p(\delta(v_i)=s))
\end{split}
\label{eqn:maHetDW}
\raisetag{35pt}
\end{equation}
where $\mathcal{T}_{\mathcal{V}}$ denotes the set of node types (e.g., author, paper, and venue), and $\mathcal{N}_t(v_i)$ denotes node $v_i$'s context nodes with type $t$. 

In order to apply~\proposed~to a HetNet, we need to consider the following two cases and model them differently:
1) case 1: the aspect of the target node is unknown, thus should be inferred, and 2) case 2: the aspect of the target node is already revealed, thus there is no need to infer its aspect. 
Concretely, taking a sample of random walk guided by meta-path APA (Figure~\ref{fig:metapath}), case 1 is when the target node is author, and case 2 is when the target node is paper.
Specifically, in case 2 shown in Figure~\ref{fig:metapath}b, we argue that the aspect (i.e. topic) of a paper can be easily inferred by looking at its content (i.e., text), whereas in case 1 shown in Figure~\ref{fig:metapath}a, the aspect of an author should still be inferred from its context. Therefore, case 1 should be modeled by our~\proposedhet~framework in Eqn.~\ref{eqn:maHetDW}, and case 2 can be simply modeled by the previous single aspect embedding method, such as metapath2vec~\cite{dong2017metapath2vec} as follows:
\begin{equation}
\small
\mathcal{J}^{(\mathbf{w})}_{\textsf{HetDW}}=\sum_{v_i \in \mathbf{w}} \sum_{t \in \mathcal{T}_{\mathcal{V}}} \sum_{v_{j} \in \mathcal{N}_{t}(v_i)}\log p(v_j | v_i)
\label{eqn:HetDW}
\end{equation}
\begin{table*}[th]
	\centering
	\small
	\caption{The overall performance for link prediction in terms of AUC-ROC (OOM: Out of memory).}
	\vspace{-2ex}
	\renewcommand{\arraystretch}{0.82}  
	\begin{tabular}{>{\centering\arraybackslash}p{1.4cm}||>{\centering\arraybackslash}p{0.7cm}>{\centering\arraybackslash}p{0.5cm}>{\centering\arraybackslash}p{0.8cm}>{\centering\arraybackslash}p{0.85cm}>{\centering\arraybackslash}p{0.8cm}|>{\centering\arraybackslash}p{0.7cm}>{\centering\arraybackslash}p{0.5cm}>{\centering\arraybackslash}p{0.8cm}>{\centering\arraybackslash}p{0.85cm}>{\centering\arraybackslash}p{0.8cm}|>{\centering\arraybackslash}p{0.7cm}>{\centering\arraybackslash}p{0.7cm}>{\centering\arraybackslash}p{0.8cm}>{\centering\arraybackslash}p{0.85cm}>{\centering\arraybackslash}p{0.8cm}}
		dim $(d\times K)$& \multicolumn{5}{c|}{100 $(d=20, K=5)$}      & \multicolumn{5}{c|}{200 $(d=40, K=5)$}      & \multicolumn{5}{c}{500 $(d=100, K=5)$} \\
		\midrule
		& DW    & DGI & PolyDW & Splitter & \proposed  & DW   & DGI & PolyDW & Splitter & \proposed  & DW    & DGI & PolyDW & Splitter & \proposed \\
		\midrule
		\midrule
		Filmtrust   & 0.6850  &0.6973 & 0.6953  & 0.6128  & \textbf{0.7426}  & 0.7399 & 0.7094& 0.6841  & 0.6111  & \textbf{0.7460}  & 0.7415 & 0.7215 & 0.6643  & 0.6097  & \textbf{0.7501}  \\
		Wiki-vote   & 0.6273  & 0.5860 & 0.5557  & 0.5190  & \textbf{0.6478}  & 0.6277 &0.5741 & 0.5179  & 0.5085  & \textbf{0.6464}  & 0.6260 & \textbf{0.6540}  & 0.5161  & 0.5048  & 0.6507  \\
		CiaoDVD     & 0.7136  & 0.6809& 0.6528  & 0.5978  & \textbf{0.7430}  & 0.7014 & 0.6696 & 0.6263  & 0.5881  & \textbf{0.7447}  & 0.7140 & 0.6897 & 0.6058  & 0.5819  & \textbf{0.7450}  \\
		BlogCatalog & 0.8734  & 0.9191 & 0.7505  & 0.8441  & \textbf{0.9503}  & {0.9220} & 0.9083 & 0.6944  & 0.8199  & \textbf{0.9548}  & 0.9331 & OOM  & 0.6249  & 0.7876  &\textbf{0.9429}  \\
		Epinions    & 0.7188  & 0.6684& 0.7038  & 0.6880  & \textbf{0.7416}  & 0.7223 & 0.6711 & 0.6884  & 0.6733  & \textbf{0.7441}  & 0.7312 & OOM  & 0.6720  &    0.6581   & \textbf{0.7459}  \\
		Flickr      & 0.9506  &0.9214 & 0.9146  &   0.9528      &  \textbf{0.9584} & 0.9580 & OOM  &      0.8862   &    0.8582     &  0.9571 & 0.9570 & OOM  &    0.8582   &   0.9299   &  \textbf{0.9678} \\
		\hline
		PPI         & 0.8236  & 0.8087& 0.7286  & 0.8372  & \textbf{0.8887}  & 0.8237 & 0.8341& 0.6995  & 0.8346  & \textbf{0.8947}  & 0.8214 & 0.8593& 0.6693  & 0.8336  & \textbf{0.8991} \\
		\hline
		Wikipedia   & 0.7729  &0.8984& 0.6259  & 0.6897  & \textbf{0.9049}  & 0.8677 & 0.8927 & 0.5920  & 0.6939  & \textbf{0.9040}  & 0.8414 & \textbf{0.9029} & 0.5218  & 0.7018  & {0.9011}  \\
		\hline
		Cora        & \textbf{0.9181}  &0.8223 & 0.8504  & 0.8357  & 0.8814  & \textbf{0.9110} & 0.8300& 0.8416  & 0.8361  & {0.9056}  & 0.8814 &\textbf{0.9475} & 0.8393  & 0.8412  & 0.9181  \\
		ca-HepTh    & \textbf{0.9080}  & 0.8661 & 0.8806  & 0.8827  & 0.8989  & \textbf{0.9160} & 0.8787 & 0.8812  & 0.9076  & 0.9119  & \textbf{0.9219} & 0.7402 & 0.8831  & 0.9058  & 0.9185  \\
		ca-AstroPh  & \textbf{0.9784}  &0.9144 & 0.9661  & 0.9731  & 0.9734  & 0.9803 & 0.9690 & 0.9734  & 0.9791  & \textbf{0.9821}  & 0.9775 & OOM & 0.9754  & 0.9827  & \textbf{0.9842 } \\
		4area       & \textbf{0.9548}  &0.9253 & 0.9441  & 0.9355  & 0.9503  & 0.9551 & 0.9349 & 0.9449  & 0.9496  & \textbf{0.9587}  & 0.9553 & OOM & 0.9463  & 0.9550  &  \textbf{0.9627} \\
	\end{tabular}%
	\label{tab:overall}%
	\vspace{-3ex}
\end{table*}%

The final objective function is to jointly optimize Eqn.~\ref{eqn:maHetDW} (case 1) and Eqn.~\ref{eqn:HetDW} (case 2) in addition to the aspect regularization framework in Eqn.~\ref{eqn:reg} as follows:
\begin{equation}
\small
\mathcal{L}_{\textsf{Het}} =-\sum_{\mathcal{P} \in \mathcal{S}(\mathcal{P})}\sum_{\mathbf{w} \in \mathcal{W}_{\mathcal{P}}} \left[\mathcal{J}^{(\mathbf{w})}_{\proposedhet} + \mathcal{J}^{(\mathbf{w})}_{\textsf{HetDW}}\right] + \lambda  \operatorname{reg}_{\textsf{asp}}
\end{equation}
where $\mathcal{S}(\mathcal{P})$ denotes all predefined meta-path schemes.

Note that although we specifically described the scenario in the author identification for the ease of explanations,~\proposedhet~can be easily applied to other tasks such as recommendation, where the task is to infer the user-item pairwise relationship.

\subsubsection{\textbf{Determining node aspect in a HetNet}. } 
\label{sec:aspect_het}
Recall that in Sec.~\ref{sec:gumbel}, given a homogeneous network, we determine the aspect of a target node based on its context nodes. In a HetNet, the node aspects can be similarly determined, however, we should take into account the heterogeneity of the node types among the context nodes. In other words, some node types might be more helpful than others in terms of determining the aspect of the target node. For example, for a target node ``author'' in Figure~\ref{fig:metapath}, a context node whose type is ``paper'' is more helpful than another author-typed node because a paper is usually written about a single topic, whereas an author can work on various research areas.
Moreover, it is important to note that determining the aspect of the target node becomes even more straightforward if node attribute information is provided, for example, the content of the paper.
In this regard, we leverage the paper abstract information to obtain the paper embedding vector, and rely only on the paper nodes to determine the aspect of the target author node. Specifically, we employ a GRU-based paper encoder introduced in~\cite{zhang2018camel} that takes pretrained word vectors of an abstract, and returns a single vector that encodes the paper content. Refer to Section 3.1 of~\cite{zhang2018camel} for more detail.

\begin{table}[]
	\caption{Statistics of the datasets. (Dir.: directed graph.)}
	\vspace{-2ex}
	\label{tab:stat}
	\renewcommand{\arraystretch}{0.95}
	\small
	\begin{tabular}{c|c|c|c|c}
		\multicolumn{3}{c|}{Dataset}                                                            & Num. nodes   & Num. edges   \\
		\hline
		\hline
		\multirow{12}{*}{{\rotatebox[origin=c]{90}{\parbox[c]{2cm}{\centering Homogeneous Network}}}} & \multirow{6}{*}{{{\rotatebox[origin=c]{90}{\parbox[c]{1.5cm}{\centering Social Network}}}}}   & Filmtrust (Dir.)    & 1,642       & 1,853       \\
		&                                   & Wiki-vote (Dir.)   & 7,066        & 103,689     \\
		&                                   & CiaoDVD (Dir.)     & 7,375       & 111,781     \\
		&                                   & BlogCatalog  & 10,312      & 333,983     \\
		&                                   & Epinions (Dir.)    & 49,290      & 487,181     \\
		&                                   & Flickr       & 80,513      & 5,899,882   \\
		\cline{2-5}
		& \multicolumn{2}{c|}{PPI}                          & 3,890        & 76,584       \\
		\cline{2-5}
		& \multicolumn{2}{c|}{Wikipedia (Word co-occurrence)}                    & 4,777        & 184,812     \\
		\cline{2-5}
				& \multirow{4}{*}{{{\rotatebox[origin=c]{90}{\parbox[c]{1.3cm}{\centering \footnotesize{Academic Network}}}}}} & Cora        & 2,708       & 5,429       \\
		&                                   & ca-HepTh     & 9,877       & 25,998      \\
		&                                   & ca-AstroPh   & 18,772      & 198,110     \\
		&                                   & 4area        & 27,199      & 66,832      \\
		
		\hline\hline
		\multirow{2}{*}{HetNet}               & \multirow{2}{*}{DBLP}             & Num. authors & Num. papers & Num. venues \\\cline{3-5}
		&                                   & 27,920       & 21,808      & 18         
	\end{tabular}
\vspace{-5ex}
\end{table}

\section{Experiments}
\noindent\textbf{Datasets. } We evaluate our proposed framework on \textbf{thirteen} commonly used publicly available datasets including a HetNet. The datasets can be broadly categorized into social network, protein network, word-co-occurrence network, and academic network. Table~\ref{tab:stat} shows the statistics of the datasets used in our experiments. Refer to the appendix for more detail.

\smallskip
\noindent\textbf{Methods Compared. }
As~\proposed~is an unsupervised multi-aspect network embedding framework that can be applied on both homogeneous and heterogeneous networks, we compare with the following unsupervised methods:
\vspace{-1ex}
\begin{enumerate}[leftmargin=0.5cm]
	\item Unsupervised embedding methods for homogeneous networks
	\begin{itemize}[leftmargin=0.1cm]
		\item Models a single aspect
		\begin{itemize}[leftmargin=0.1cm]
			\item \textbf{\textbf{Deepwalk}}~\cite{perozzi2014deepwalk}/\textbf{\textbf{node2vec}}~\cite{grover2016node2vec}:  They learn node embeddings by random walks and skip-gram. As they generally show similar performance, we report the best performing method among them.
			\item
			\textbf{DGI}~\cite{velivckovic2018deep}: It is the state-of-the-art unsupervised network embedding method that maximizes the mutual information between the graph summary and the local patches of a graph.
		\end{itemize}
		\item Models multiple aspects
		\begin{itemize}[leftmargin=0.1cm]
			\item \textbf{PolyDW}~\cite{liu2019single}: It performs MF--based clustering to obtain aspect distribution for each node from which an aspect for the target and the context nodes are independently sampled.
			\item \textbf{Splitter}~\cite{epasto2019single}: It splits each node into multiple embeddings by performing local graph clustering, called persona2vec.
		\end{itemize}
	\end{itemize}
	\item Unsupervised embedding methods for heterogeneous networks
	\begin{itemize}[leftmargin=0.1cm]
		\item Models a single aspect
		\begin{itemize}[leftmargin=0.1cm]
			\item \textbf{M2V++}~\cite{dong2017metapath2vec}: It learns embeddings for nodes in a HetNet by performing meta-path guided random walk followed by heterogeneous skip-gram. Following~\cite{park2019task}, we leverage the paper abstract for paper embeddings.
			\item\textbf{Camel}~\cite{zhang2018camel}: It is a task-guided heterogeneous network embedding method developed for the author identification task in which content-aware skip-gram is introduced.
		\end{itemize}
		\item Models multiple aspects
		\begin{itemize}[leftmargin=0.1cm]
		\item \textbf{TaPEm}~\cite{park2019task}: It is the state-of-the-art task-guided heterogeneous network embedding method that introduces the pair embedding framework to directly capture the pairwise relationship between two heterogeneous nodes.
		\end{itemize}
	\end{itemize}
\end{enumerate}

\noindent\textbf{Experimental Settings and Evaluation Metrics. }
For evaluations of~\proposed~and~\proposedhet, we perform link prediction and author identification, respectively. For link prediction, we follow the protocol of  Splitter~\cite{epasto2019single,abu2017learning}, 
and for author identification, we follow the protocol of~\cite{zhang2018camel,park2019task}. As for the evaluation metrics, we use AUC-ROC for link prediction, 
and recall, F1, and AUC for author identification. 
Note that for all the tables in this section, the number of aspects $K$ is 1 for all the single aspect embedding methods, i.e. DW, DGI, M2V++, Camel, and TaPEm.
For more information regarding the hyperparameters and the model training--related details, refer to the appendix.

\subsection{Performance Analysis}
\noindent\textbf{Overall Evaluation. } 
Table~\ref{tab:overall} shows the link prediction performance of various methods. We have the following observations: 
\textbf{1)}~\proposed~generally performs well on all datasets, especially outperforming other methods on social networks, PPI and Wikipedia networks.
We attribute such behavior to the fact that nodes in these networks inherently exhibit multiple aspects compared with nodes in the academic networks, where each node is either a paper (Cora) or an author (ca-HepTh, ca-AstroPh, and 4area) that generally focuses on a single research topic, thus we have less distinct aspects to be captured. In particular, nodes (authors) in ca-AstroPh and ca-HepTh networks are even more focused as they are specifically devoted to Astro Physics and High Energy Physics Theory, respectively. On the other hand, social networks contain various communities, for example, BlogCatalog is reported to have 39 ground truth groups in the network. 
\textbf{2)} We observe that the prediction performance on the academic networks are generally high for all methods, which aligns with the results reported in \cite{epasto2019single} where DW generally performs on par with Splitter under the same embedding size. We conjecture that this is because links in academic networks are relatively easy to predict because academic research communities are relatively focused, thereby nodes having less diverse aspects as mentioned above. 
\textbf{3)} We observe that DGI generally performs better as the embedding size increases, outperforming others in some datasets when $d=500$. However, DGI is not scalable to large dimension sizes, which is also mentioned in the original paper~\cite{velivckovic2018deep} (DGI fixed $d=512$ for all datasets, but due to memory limitation, $d$ is reduced to 256 for pubmed dataset that contains 19,717 nodes). 

\begin{table}[t]
	\centering
	\small
	\renewcommand{\arraystretch}{0.8}
	\caption{Benefit of the Gumbel-Softmax. }
	\vspace{-2ex}
	\begin{tabular}{c||c|c||c}
		$d=20, K=5$ & Softmax & Gumbel-Softmax & Improvement\\
		\hline
		\hline
		Filmtrust & 0.6421 & \textbf{0.7426} &15.65\%\\
		Wiki-vote & 0.6165 & \textbf{0.6478}&  5.08\%\\
		CiaoDVD & 0.6162 & \textbf{0.7430} &20.58\%\\
		BlogCatalog & 0.7323 & \textbf{0.9503} & 29.77\%\\
		Epinions & 0.6693 & \textbf{0.7416} &10.80\%\\
		Flickr &   0.8956    & \textbf{0.9584} &7.01\%\\
		\hline
		PPI   & 0.6919 & \textbf{0.8887} &28.44\%\\
		\hline
		Wikipedia & 0.8269 & \textbf{0.9049}& 9.43\%\\
		\hline
		Cora  & {0.8605} & \textbf{0.8814} &2.43\%\\
		ca-HepTh & 0.8890 & \textbf{0.8989}& 1.11\%\\
		ca-AstroPh & 0.9116 & \textbf{0.9734} &6.78\%\\
		4area &   0.9286    & \textbf{0.9503} &2.34\%\\
	\end{tabular}%
	\label{tab:gumbel}%
	\vspace{-2ex}
\end{table}%

\begin{figure}[t]
	\centering
	\includegraphics[width=0.85\linewidth]{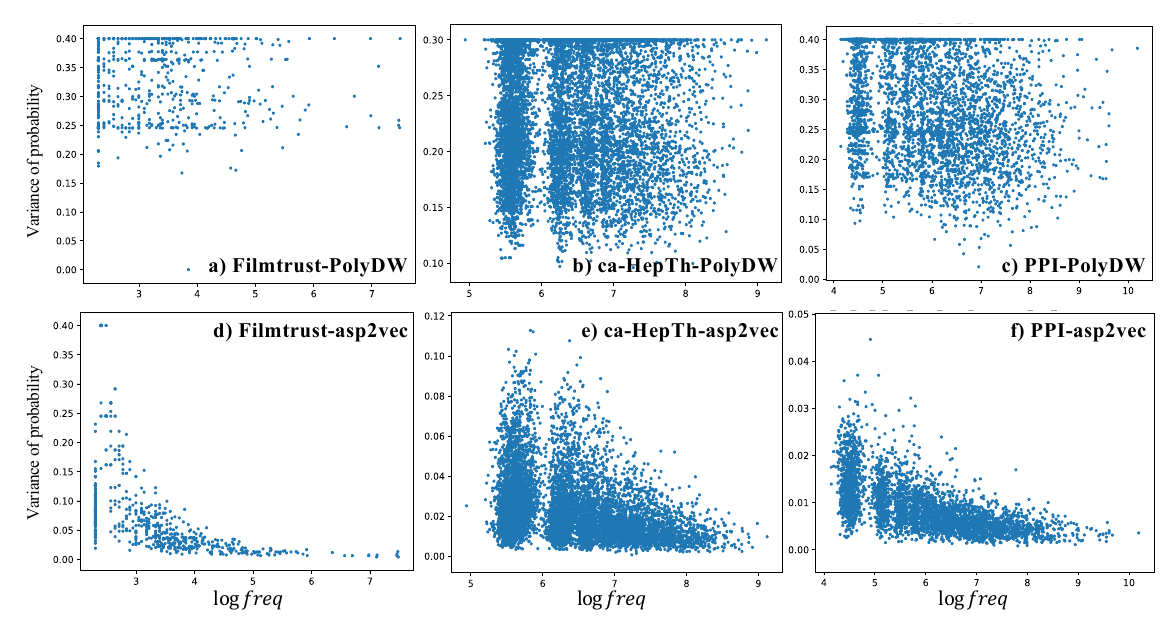}
	\vspace{-3ex}
	\caption{Variance of the aspect selection probability over the frequency of the node appearing in random walks (Top row: PolyDW, Bottom row:~\proposed).}
	\label{fig:variance}	
	\vspace{-4ex}
\end{figure}

\smallskip
\noindent\textbf{Benefit of Aspect Selection Module. }
Table~\ref{tab:gumbel} shows the comparisons between~\proposed~with conventional softmax (Eqn.~\ref{eqn:SM}), and the Gumbel-Softmax trick (Eqn.~\ref{eqn:gumbel}). We found that the Gumbel-Softmax indeed is beneficial, and more importantly, we observe that the improvements are more significant for social networks, PPI, and Wikipedia network, compared with the academic networks. 
This verifies that the aspect modeling is more effective for networks with inherently diverse aspects, such as social and PPI networks.

To further verify the benefit of the aspect selection module compared with the offline clustering--based method, we examine how the aspects are actually assigned to nodes.
Our assumption is nodes that frequently appear in random walks (i.e., $\mathcal{W}$) are more likely to exhibit more aspects compared with the less frequently appearing ones. Therefore, the variance of aspect probability distribution of a frequently appearing node will be relatively smaller than that of the less frequently appearing one. For example, given four aspects, a frequently appearing node may have an aspect probability distribution [0.2,0.3,0.3,0.2], whereas that of a less frequently appearing node is likely to have a skewed distribution that looks like [0.7,0.0,0.3,0.0] with a higher variance. In Figure~\ref{fig:variance}, we plot the variance of the aspect probability distribution of each target node according to its frequency within the random walks. Precisely, each dot in the figure represents the variance of the aspect distribution to which each target node is assigned.
As expected,~\proposed~(Figure~\ref{fig:variance} bottom) tends to assign low variance aspect distribution to high frequency nodes, and low frequency nodes have high variance aspect distribution. On the other hand, as PolyDW determines aspects by an offline clustering algorithm, there is no such tendency observed (Figure~\ref{fig:variance} Top), which verifies the superiority of our context--based aspect selection module.

\begin{table}[t]
	\centering
	\small
	\caption{Link prediction performance (AUC-ROC) without $\textsf{reg}_{\textsf{asp}}$, and over various thresholds ($\epsilon$).}
	\vspace{-3ex}
	\renewcommand{\arraystretch}{0.8}
	\begin{tabular}{>{\centering\arraybackslash}p{1.6cm}|>{\centering\arraybackslash}p{0.45cm}|>{\centering\arraybackslash}p{0.45cm}>{\centering\arraybackslash}p{0.45cm}>{\centering\arraybackslash}p{0.45cm}>{\centering\arraybackslash}p{0.45cm}>{\centering\arraybackslash}p{0.5cm}|c}
		\multicolumn{1}{c|}{\multirow{2}[1]{*}{\begin{tabular}[x]{@{}c@{}}dim $=100$\\($d=20, K=5$)\end{tabular}}} & \multicolumn{1}{c|}{\multirow{2}[0]{*}{\begin{tabular}[x]{@{}c@{}}No  \\$\operatorname{reg}_{\textsf{asp}}$\end{tabular}}} & \multicolumn{5}{c|}{Threshold $(\epsilon)$} & \multicolumn{1}{c}{\multirow{2}[0]{*}{\begin{tabular}[x]{@{}c@{}} best vs.  \\ No $\textsf{reg}_{\textsf{asp}}$\end{tabular}}} \\
		\cline{3-7}          &       & \multicolumn{1}{c}{0.9} & \multicolumn{1}{c}{0.7} & \multicolumn{1}{c}{0.5} & \multicolumn{1}{c}{0.3} & \multicolumn{1}{c|}{0.1} &  \\
		\hline
		\hline
		Filmtrust & 0.660  & \textbf{0.743 } & 0.742  & 0.740  & 0.738  & 0.735  & 12.58\% \\
		Wiki-vote & 0.616  & 0.647  & \textbf{0.648 } & 0.647  & 0.647  & 0.645  & 5.15\% \\
		CiaoDVD & 0.617  & \textbf{0.743 } & 0.742  & 0.742  & 0.738  & 0.735  & 20.37\% \\
		BlogCatalog & 0.791  & 0.948  & \textbf{0.950 } & 0.949  & 0.939  & 0.869  & 20.11\% \\
		Epinions & 0.684  & \textbf{0.742 } & 0.741  & 0.738  & 0.731  & 0.693  & 8.37\% \\
		Flickr & 0.897  & 0.955  & \textbf{0.958 } & 0.954  & 0.954  & 0.929  & 6.85\% \\
		\hline
		PPI   & 0.729  & 0.880  & 0.885  & \textbf{0.889 } & 0.881  & 0.819  & 21.97\% \\
		\hline
		Wikipedia & 0.841  & 0.896  & 0.904  & \textbf{0.905 } & 0.880  & 0.850  & 7.60\% \\
		\hline
		Cora  & 0.879  & 0.881  & 0.880  & \textbf{0.881} & 0.862  & 0.857  & 0.23\% \\
		ca-HepTh & 0.879  & \textbf{0.899 } & 0.896  & 0.898  & 0.893  & 0.864  & 2.30\% \\
		ca-AstroPh & 0.921  & \textbf{0.973} & 0.973  & 0.971  & 0.967  & 0.939  & 5.56\% \\
		4area & 0.919  & \textbf{0.950 } & 0.949  & 0.946  & 0.940  & 0.915  & 3.44\% \\
	\end{tabular}%
	\vspace{-2ex}
	\label{tab:epsilon}%
\end{table}%

\begin{figure}[h]
	\centering
	\includegraphics[width=0.85\linewidth]{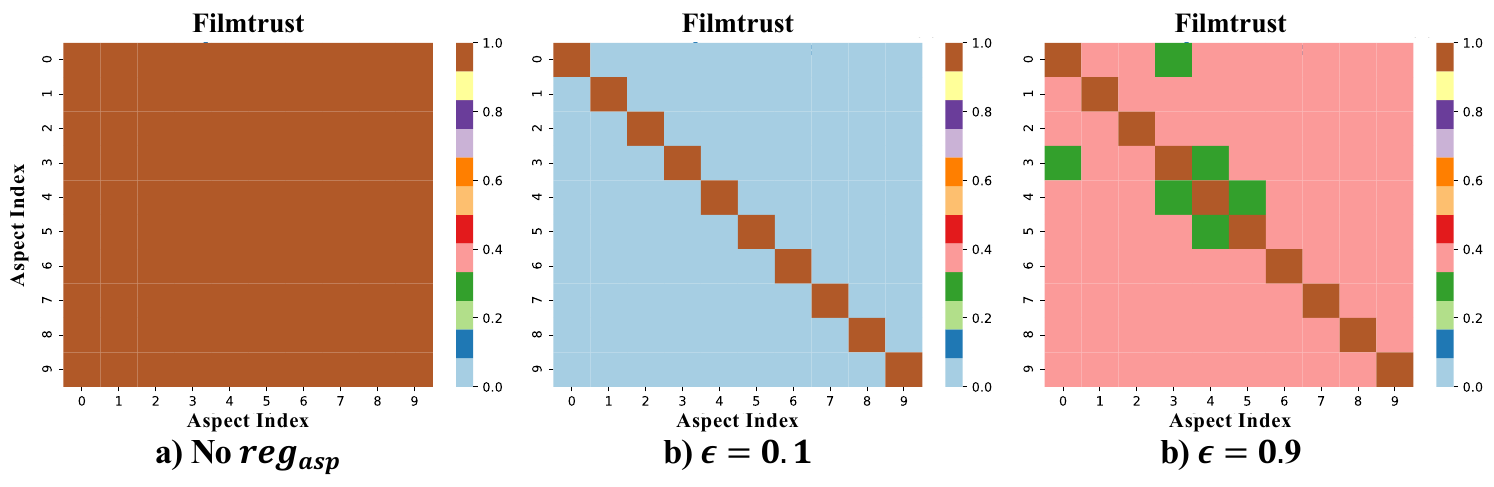}
	\vspace{-2ex}
	\caption{Heatmaps showing the similarity between the aspect embeddings (Best viewed in color). }
	\label{fig:heatmap}	
	\vspace{-2ex}
\end{figure}
\smallskip
\noindent\textbf{Benefit of Aspect Regularizer. }
Table~\ref{tab:epsilon} shows the link prediction AUC-ROC over various thresholds ($\epsilon$) for $\textsf{reg}_{\textsf{asp}}$. 
\textbf{1)} We observe that the performance drops significantly when $\epsilon=1$, that is, when the aspect regularization framework is not incorporated (no $\textsf{reg}_\textsf{asp}$), which verifies the benefit of the aspect regularization framework.
\textbf{2)} The aspect regularization framework is less effective on the academic networks (see best vs. No $\textsf{reg}_{\textsf{asp}}$). Again, this is because academic networks inherently have less diverse aspects, which reduce the need for modeling diverse aspect embeddings.
\textbf{3)} In Figure~\ref{fig:heatmap}, we illustrate heatmaps where each element denotes the cosine similarity between a pair of aspect embedding vectors. More precisely, we compute the pairwise similarity between aspect embedding vectors for all nodes, and then compute their average to obtain the final heatmap. We observe that the aspect embeddings are trained to be highly similar to each other without $\textsf{reg}_{\textsf{asp}}$ (Figure~\ref{fig:heatmap}a), which shows the necessity of the aspect regularization framework. Moreover, when $\epsilon=0.1$ (Figure~\ref{fig:heatmap}b), the similarity among the aspect embeddings is small, which demonstrates that a small $\epsilon$ encourages the aspect embeddings to be diverse. Finally, when $\epsilon=0.9$ (Figure~\ref{fig:heatmap}c), we observe that some of the aspect embeddings are trained to be less similar (i.e. green) than others (i.e. red) allowing more flexibility in learning the aspect embeddings.

\begin{table}[htbp]
	\centering
	\small
	\caption{Author identification results in terms of recall, F1, and AUC over various dimensions ($d\times K$).}
	\vspace{-3ex}
	\renewcommand{\arraystretch}{0.6}
	\begin{tabular}{c|c||cc||cc||c}
		\multirow{2}[1]{*}{\begin{tabular}[x]{@{}c@{}}dim\end{tabular}} & \multirow{2}[1]{*}{Methods} & \multicolumn{2}{c||}{Recall@N} & \multicolumn{2}{c||}{F1@N} & \multirow{2}[1]{*}{AUC} \\
		\cline{3-6}
		&       & 5     & 10    & 5     & 10    &  \\
		\hline
		\hline
		\multirow{3}[2]{*}{200} & M2V++ & 0.4607  & 0.6115  & 0.2691  & 0.2100  & 0.8752  \\
		& Camel & 0.5774  & 0.7270  & 0.3384  & 0.2518  & 0.9021  \\
		& TaPEm & 0.6150  & 0.7473  & 0.3593  & 0.2586  & 0.9031  \\
		& \proposedhet &   \textbf{0.6474}    &   \textbf{0.7701}    &   \textbf{0.3832}   &    \textbf{0.2709}   & \textbf{0.9137} \\
		\hline
		\multirow{3}[2]{*}{500} & M2V++ & 0.4137  & 0.5662  & 0.2442  & 0.1970  & 0.8433  \\
		& Camel & 0.4788  & 0.6243  & 0.2827  & 0.2197  & 0.8602  \\
		& TaPEm & 0.5193  & 0.6592  & 0.3058  & 0.2319  & 0.8643  \\
		& \proposedhet &   \textbf{0.6601}   &   \textbf{0.7671}    &    \textbf{0.3879}   &    \textbf{0.2693}   &  \textbf{0.9161}
	\end{tabular}%
	\label{tab:het}%
	\vspace{-3ex}
\end{table}%

\smallskip
\noindent\textbf{Evaluations on HetNet. }
Table~\ref{tab:het} shows the result of author identification. 
\textbf{1)} We observe that~\proposedhet~significantly outperforms the state-of-the-art task-guided author identification methods in all the metrics, which verifies the benefit of our context--based aspect embedding framework. More precisely,~\proposedhet~outperforms TaPEm~\cite{park2019task}, which is a recently proposed method whose motivation is similar to ours in that it models multiple aspects of each node by the pair embedding framework. This verifies that explicitly learning multiple aspect embedding vectors for each node is superior to learning an embedding vector for every possible node pairs.
\textbf{2)} Recall that there are multiple types of nodes in a HetNet, and some types of nodes are more helpful than others in determining the aspect of a target node. Table~\ref{tab:het_aspect} shows the performance when different types of context nodes are used for determining the aspect of the target node, i.e., author node. We observe that 
using paper (type) only performs the best, whereas using author only performs the worst.
This is expected because we encode the paper abstract to generate paper embeddings, and thus the aspect of the target author can be readily determined by examining paper nodes within its context nodes. On the other hand, the author embeddings only capture the local structural information, which is less deterministic of the target aspect compared to the node attributes.

\begin{table}[htbp]
	\centering
	\small
	\vspace{-2ex}
	\caption{Using different types of context for~\proposedhet.}
	\vspace{-3ex}
	\renewcommand{\arraystretch}{0.7}
	\begin{tabular}{c|c||c|c|c}
		\hline
		\multicolumn{5}{c}{\proposedhet} \\
		\midrule
		\multicolumn{2}{c||}{dim$ = 200\ (d=40, K = 5)$} & {Recall@10} & {F1@10} & AUC \\
		\midrule
		\midrule
		\multicolumn{1}{c|}{\multirow{3}[1]{*}{Context Type}} & Author & 0.7627 & 0.2677 & 0.9099 \\
		& Paper & \textbf{0.7701} & \textbf{0.2709} & \textbf{0.9137} \\
		& Author+Paper & 0.7682 & 0.2695 & 0.9127 \\
	\end{tabular}%
	\label{tab:het_aspect}%
	\vspace{-3ex}
\end{table}%

\section{Conclusion}
In this paper, we present a novel multi-aspect network embedding method called~\proposed~that dynamically determines the aspect based on the context information. Major components of~\proposed~ is 1) the aspect selection module, which is based on the Gumbel-Softmax trick to approximate the discrete sampling of the aspects, and facilitate end-to-end training of the model, and 2) the aspect regularization framework that encourages the learned aspect embeddings to be diverse, but to some extent related to each other. We also demonstrate how to extend~\proposed~to a HetNet. Through experiments on multiple networks with various types, we empirically show the superiority of our proposed framework. 

\smallskip
\noindent\textbf{Acknowledgment}: IITP2018-0-00584, IITP2019-0-01906, IITP2019-2011-1-00783, 2017M3C4A7063570, 2016R1E1A1A01942642
\vspace{-2ex}
\bibliographystyle{ACM-Reference-Format}
\bibliography{acmart}

\appendix
\section*{Appendix for Reproducibility}
\section{Experimental Details}
\smallskip
\subsection{Code and Datasets}
\proposed~is implemented using PyTorch. The source code and instructions on how to run the code can be found here\footnote{https://github.com/pcy1302/asp2vec}.
Table~\ref{tab:codeurl} shows the urls of the author implementations of the compared methods, and Table~\ref{tab:dataurl} shows the urls from which the datasets can be downloaded. Nodes in social networks refer to users, nodes in PPI network refer to protein, nodes Wikipedia refer to words, nodes in Cora (citation network) refer to papers, and nodes in ca-HepTh, ca-AstroPh and 4area (all three are co-authorship networks) refer to authors.
Each node in Cora is associated with a bag-of-words representation of a document (size=1433), whereas the remaining datasets do not have node attribute information.
Finally, as DGI uses node attribute information, we used the adjacency matrix to initialize the node attribute matrix to apply DGI on datasets without node attribute information (i.e., all but Cora).
\begin{table}[h]
	\centering
	\small
	\caption{URL links to the code.}
	\vspace{-3ex}
	\begin{tabular}{l||l}
		Methods & URL link to the code \\
		\midrule
		\midrule
		DGI   & https://github.com/PetarV-/DGI \\
		\hline
		PolyDW & github.com/ninghaohello/Polysemous-Network-Embedding \\
		\hline
		Splitter & 
		\begin{tabular}[x]{@{}l@{}}github.com/google-research/\\google-research/tree/master/graph\_embedding/persona\end{tabular}
		\\
		\hline
		camel & https://github.com/chuxuzhang/WWW2018\_Camel \\
		\hline
		TaPEm & https://github.com/pcy1302/TapEM \\
	\end{tabular}%
	\label{tab:codeurl}%
\end{table}%

\begin{table}[h]
	\centering
	\small
	\caption{URL links to the datasets.}
	\begin{tabular}{l|l}
		& \multicolumn{1}{c}{URL link to the dataset} \\
		\midrule
		\midrule
		Filmtrust & https://www.librec.net/datasets.html\#filmtrust \\
		Cora  & https://github.com/tkipf/pygcn/tree/master/data/cora \\
		PPI   & https://snap.stanford.edu/node2vec/ \\
		Wikipedia & https://snap.stanford.edu/node2vec/ \\
		Wiki-vote & https://snap.stanford.edu/data/index.html \\
		CiaoDVD & https://www.librec.net/datasets.html\#filmtrust \\
		ca-HepTh & https://snap.stanford.edu/data/index.html \\
		BlogCatalog & http://socialcomputing.asu.edu/datasets/BlogCatalog3 \\
		ca-AstroPh & https://snap.stanford.edu/data/index.html \\
		4area & http://www.perozzi.net/projects/focused-clustering/ \\
		Epinions & http://www.trustlet.org/downloaded\_epinions.html \\
		Flickr & http://socialcomputing.asu.edu/datasets/Flickr \\
		\midrule
		DBLP  & https://github.com/pcy1302/TapEM/tree/master/data \\
	\end{tabular}%
	\label{tab:dataurl}%
\end{table}%

\subsection{Evaluation Protocol}
\noindent\textbf{Homogeneous Network: Link Prediction. }
For link prediction in homogeneous networks, we follow the convention of~\cite{abu2017learning,grover2016node2vec,epasto2019single}. More precisely, we first split the original graph into to two equal sized set of edges, i.e., $\text{E}_{\text{train}}$, and $\text{E}_{\text{test}}$. To obtain $\text{E}_{\text{test}}$, we randomly remove the edges while preserving the connectivity of the original graph. 
To obtain the negative counterparts of $\text{E}_{\text{train}}$ and $\text{E}_{\text{test}}$, we randomly generate a set of edges of size $2\times|\text{E}_{\text{test}}|$ and split it in half into  $\text{E}_{\text{train}_\text{neg}}$ and $\text{E}_{\text{test}_\text{neg}}$, respectively. We train a logistic regression classifier on the $\text{E}_{\text{train}}\cup\text{E}_{\text{train}_\text{neg}}$ ~\cite{epasto2019single,abu2017learning}, and the link prediction performance is then measured by ranking (AUC-ROC) the removed edges $\text{E}_{\text{test}}$ among $\text{E}_{\text{test}}\cup\text{E}_{\text{test}_\text{neg}}$. Finally, for the above preprocessing, we used~\footnote{https://github.com/google/asymproj\_edge\_dnn/blob/master/create\_dataset\_arrays.py}, which is from the implementation of~\cite{abu2017learning}.

\smallskip
\noindent\textbf{Heterogeneous Network: Author Identification. }
For author identification in a heterogeneous network, we follow the convention of~\cite{zhang2018camel,park2019task}. More precisely, we use papers published before timestamp T ($=2013$) for training, and split papers that are published after timestamp T in half to make validation and test datasets. We report the test performance when the validation result is the best. For final evaulation, we randomly sample a set of negative authors and combine it with the set of true authors to generate 100 candidate authors for each paper, and then rank the positive paper among them.
\begin{table}[htbp]
	\centering
	\small
	\caption{Link prediction result over various $d$ and $K$ under a limited $d\times K$.}
	\vspace{-2ex}
	\renewcommand{\arraystretch}{0.7}
	\begin{tabular}{c||c|c|c|c}
		& \multicolumn{4}{c} {dim$=200 \ (d \times K = 200)$} \\
		\cmidrule{2-5}          & \begin{tabular}[x]{@{}c@{}}$d=20,$\\$K=10$\end{tabular} & \begin{tabular}[x]{@{}c@{}}$d=50,$\\$K=4$\end{tabular} & \begin{tabular}[x]{@{}c@{}}$d=40,$\\$K=5$\end{tabular} & \begin{tabular}[x]{@{}c@{}}$d=100,$\\$K=2$\end{tabular} \\
		\hline
		\hline
		Filmtrust & 0.7426  & 0.7455  & 0.7425  & 0.7537  \\
		Wiki-vote & 0.6478  & 0.6450  & 0.6480  & 0.6476  \\
		CiaoDVD & 0.7430  & 0.7445  & 0.7385  & 0.7370  \\
		BlogCatalog & 0.9503  &   0.9538    &   0.9515    & 0.9292 \\
		Epinions &    0.7416   &   0.7436    &    0.7441   & 0.7370\\
		Flickr &    0.9584   &    0.9524   &  0.9571     & 0.9629  \\
		\hline
		PPI   & 0.8887  & 0.8947  & 0.8947  & 0.8530  \\
		\hline
		Wikipedia & 0.9049  & 0.9040  & 0.8987  & 0.8783  \\
		\hline
		Cora  & 0.8814  & 0.8827  & 0.8770  & 0.8701  \\
		ca-HepTh & 0.8989  & 0.9051  & 0.9088  & 0.9082  \\
		ca-AstroPh &  0.9734     &     0.9803  &  0.9821     &0.9812  \\
		4area &  0.9503     &  0.9547     &   0.9573    &  0.9600
	\end{tabular}%
	\label{tab:dk}%
\end{table}%

\begin{table*}[t]
	\centering
	\small
	\caption{Link prediction result of~\proposed~for various number of aspects given a fixed $d$. }
	\vspace{-3ex}
	\begin{tabular}{c||ccccccccc}
		\multirow{2}[3]{*}{\begin{tabular}[x]{@{}c@{}}Num. \\Aspects $(K)$\end{tabular}} & \multicolumn{9}{c}{dim $= d\times K$ $(d=20)$} \\
		\cmidrule{2-10}          & 2     & 3     & 4     & 5     & 6     & 7     & 8     & 9     & 10 \\
		\midrule
		\midrule
		Filmtrust & 0.7427  & 0.7408  & 0.7381  & 0.7426  & 0.7412  & 0.7398  & 0.7436  & 0.7417  & 0.7430  \\
		Wiki-vote & 0.6466  & 0.6474  & 0.6484  & 0.6478  & 0.6479  & 0.6476  & 0.6483  & 0.6483  & 0.6483  \\
		CiaoDVD & 0.7419  & 0.7428  & 0.7427  & 0.7430  & 0.7431  & 0.7436  & 0.7435  & 0.7437  & 0.7439  \\
		BlogCatalog & 0.9227  & 0.9448  & 0.9504  & 0.9503  & 0.9500  & 0.9489  & 0.9483  & 0.9479  &  0.9475\\
		Epinions & 0.7285  & 0.7389  & 0.7410  & 0.7416  & 0.7415  & 0.7414  & 0.7416  &   0.7415    & 0.7415 \\
		Flickr &    0.9267   &   0.9459    &   0.9520    &   0.9584    &   0.9519    &   0.9508    &   0.9492    &    0.9480   & 0.9476 \\
		\hline
		PPI   & 0.8531  & 0.8814  & 0.8875  & 0.8887  & 0.8891  & 0.8883  & 0.8893  & 0.8896  & 0.8897  \\
		\hline
		Wikipedia & 0.8928  & 0.8792  & 0.9022  & 0.9049  & 0.9053  & 0.9045  & 0.9034  & 0.9024  & 0.9009  \\
		\hline
		Cora  & 0.8831  & 0.8835  & 0.8872  & 0.8814  & 0.8877  & 0.8880  & 0.8869  & 0.8860  & 0.8849  \\
		ca-HepTh & 0.8977  & 0.9005  & 0.9001  & 0.8989  & 0.8978  & 0.8962  & 0.8950  & 0.8952  & 0.8955  \\
		ca-AstroPh & 0.9610  & 0.9699  & 0.9730  & 0.9734  & 0.9731  &    0.9728   &     0.9725  & 0.9721      & 0.9717 \\
		4area & 0.9496  & 0.9512  & 0.9511  & 0.9503  &    0.9491   &    0.9479   &      0.9453 &   0.9447    &  0.9438\\
		
	\end{tabular}%
	\label{tab:aspect}%
\end{table*}%

\subsection{Experimental Settings}
\begin{itemize}[leftmargin=0.1cm]
	\item For all the compared methods, we use the hyperparameters that are reported to be the best by the original papers. For PolyDW~\cite{liu2019single}, number of random walks for each node is 110, the length of each walk is 11, window size is 4, number of samples per walk ($R$ in ~\cite{liu2019single}) is 10, and the number of negative samples is 5. For Splitter~\cite{epasto2019single}, number of random walks for each node is 40, the length of each walk is 10, the window size is set to 5, and the number of negative samples is 5.
	We found that the above setting works the best.
	\item Prior to training~\proposed, we perform warm-up step. More precisely, we initialize target and aspect embedding vectors of~\proposed~with the final trained embedding vecotrs of Deepwalk~\cite{perozzi2014deepwalk}. We found that the warm-up step usually gives better performance.
	\item  For DW and~\proposed, we follow the setting of node2vec~\cite{grover2016node2vec} where the number of random walks for each node is 10, the length of each walk is 80, the window size is 3, and the number of negative samples is 2. $\tau$ and $\lambda$ of~\proposed~are set to 0.5, and 0.01, respectively.
	Note that we also tried other settings than those mentioned above, but no significant different results were obtained.
	\item Moreover, we first tried gensim implementation\footnote{https://radimrehurek.com/gensim/models/word2vec.html} of DW~\cite{perozzi2014deepwalk} / node2vec~\cite{grover2016node2vec} but we found that the results were inconsistent. Specifically, for BlogCatalog, PPI, Wikipedia, and Cora, we found the performance of gensim DW significantly underperformed our implementation of DW. Therefore, we reported the results of our DW/node2vec which produced consistent results that were consistently better than gensim implementation. 
	\item It is important to note that while we successfully reproduced the results of Splitter on ca-HepTh, and ca-AstroPh reported in~\cite{epasto2019single}, Wiki-vote could not be reproduced. This is because the implementation of Splitter is intended for an undirected graph, whereas Wiki-vote is a directed graph. We conjecture that Epasto~\etal~\cite{epasto2019single} converted Wiki-vote into an undirected graph, and reported the results. However, we used a directed version of Wiki-vote as input, which decreased the link prediction performance.
	\item For fair comparisons, we fixed the number of dimensions for each node throughout all the compared methods. For example, if $d=100$ for DW, then we set $d=20$ and $K=5$ for~\proposed~to make the total dimension size $100$, even though we do not concatenate the aspect embedding vectors.
	
	\item For~\proposedhet, we set the window size to 3, and number of negative samples to 2, and for a fair comparison, we follow~\cite{park2019task} and we use a metapath ``APA''.
\end{itemize}

\smallskip
\noindent\textbf{Other Results. }
Table~\ref{tab:aspect} shows the results for various number of aspects $K$ when $d$ is fixed to 20. We observe that $K=5$ consistently gives competitive results albeit not the best for some cases. This is the reason why we fixed $K=5$ in our experiments. Table~\ref{tab:dk} shows results for different combinations of $d$ and $K$ that multiplied to be 200. We observe that the best combination of $d$ and $K$ differs according to the datasets, where we could observe even better results compared with the results reported in our paper. Again, since $d=50$ and $K=4$ consistently gives competitive results, we used it for our experiments with dim=200 in our paper. Algorithm~\ref{pseudocode} shows the pseudocode for training~\proposed.

\begin{algorithm}
	\small
	\SetAlgoLined
	\SetKwInOut{Input}{Input}
	\SetKwInOut{Output}{Output}
	\Input{A network $\mathcal{G}=(\mathcal{V},\mathcal{E})$, target embedding matrix $\mathbf{P}\in\mathbb{R}^{|V|\times d}$, aspect embedding matrix $\{\mathbf{Q}_*^{(s)}\in\mathbb{R}^{|V|\times d}\}_{s=1}^K$
	}
	\Output{Final embedding matrix $\mathbf{U}\in\mathbb{R}^{|V|\times d}$}
	Perform a set of random walks $\mathcal{W}$ on $\mathcal{G}$
	
	\While{Convergence}{
		\ForEach{$\mathbf{w}\in\mathcal{W}$}{%
			$\mathcal{J}^{(\textbf{w})}_{\proposed}\leftarrow\textsf{run\_asp2vec}(\mathbf{w})$
		}
		$\mathcal{L}_\proposed=-\sum_{\mathbf{w}\in\mathcal{W}}\mathcal{J}_{\proposed}^{(\mathbf{w})}$  \tcp*{Eqn. 8} 
		
		$\text{reg}_{\textsf{asp}}=\sum_{i=1}^{K-1}\sum_{j=i+1}^{K}\operatorname{A-Sim}(\mathbf{Q}_*^{(i)},\mathbf{Q}_*^{(j)})$ \tcp*{Eqn. 10} 
		
		$\mathcal{L} = \mathcal{L}_{\proposed} + \lambda  \operatorname{reg}_{\textsf{asp}}$ \tcp*{Eqn. 15} 
		
		Update embedding matrices $\mathbf{P}$ and $\{\mathbf{Q}_*^{(s)}\}_{s=1}^K$ by minimizing $\mathcal{L}$
	}
	Compute final embedding matrix $\mathbf{U}$ \tcp*{Eqn. 16} 
	
	\DontPrintSemicolon
	\SetKwFunction{FMain}{\textsf{run\_asp2vec}}
	\SetKwProg{Fn}{Function}{:}{}
	\Fn{\FMain{$\mathbf{w}$}}{
		\ForEach{$v_i\in\mathbf{w}$}{%
			\ForEach{$v_j\in\mathcal{N}(v_i)$}		{
				
				\For{$s = 1;\ s < K;\ s = s + 1$}{%
					Compute $p(\delta(v_i)=s|\mathcal{N}(v_i))$ \tcp*{Eqn. 7}
					
					$p(v_j | v_i, p(\delta(v_i)=s))\leftarrow\frac{\exp(\langle \mathbf{P}_i,\mathbf{Q}^{(s)}_j\rangle)}{\sum_{v_{j^\prime}\in \mathcal{V}}\exp(\langle \mathbf{P}_i,\mathbf{Q}^{(s)}_{j{^\prime}}\rangle)}$
				}
				
				{$\log p(v_j | v_i)\leftarrow\sum_{s=1}^{K}p(\delta(v_i)=s|\mathcal{N}(v_i))\log p(v_j | v_i, p(\delta(v_i)=s))$}
			}
			$\mathcal{J}_{\proposed}^{\textbf{(w)},\mathcal{N}(v_i)}\leftarrow\sum_{v_j \in \mathcal{N}(v_i)}\log p(v_j | v_i)$
		}
		$\mathcal{J}^{(\textbf{w})}_{\proposed}\leftarrow\sum_{v_i\in\mathbf{w}}\mathcal{J}_{\proposed}^{\textbf{(w)},\mathcal{N}(v_i)}$ \tcp*{Eqn. 2}  
		
		\KwRet $\mathcal{J}^{(\textbf{w})}_{\proposed}$ \;
	}
	\caption{Pseudocode for training~\proposed.}
	\label{pseudocode}
\end{algorithm}
\vspace{-3ex}
\end{document}